\documentclass[sigconf]{acmart}
\AtBeginDocument{%
  }

\copyrightyear{2025}
\acmYear{2025}
\setcopyright{acmlicensed}\acmConference[CIKM '25]{Proceedings of the 34th ACM International Conference on Information and Knowledge Management}{November 10--14, 2025}{Seoul, Republic of Korea}
\acmBooktitle{Proceedings of the 34th ACM International Conference on Information and Knowledge Management (CIKM '25), November 10--14, 2025, Seoul, Republic of Korea}
\acmDOI{10.1145/3746252.3760929}
\acmISBN{979-8-4007-2040-6/2025/11}

\usepackage{multirow}
\usepackage{colortbl}
\usepackage{graphicx} 

\begin{document}

\title{Uncovering the Persuasive Fingerprint of LLMs in Jailbreaking Attacks}

\author{Havva Alizadeh Noughabi}
\email{havva@uoguelph.ca}
\orcid{0000-0002-9801-427X}
\affiliation{%
  \institution{Cyber Science Lab, University of Guelph}
  \city{Guelph}
  \state{ON}
  \country{Canada}
}
\author{Julien Serbanescu}
\orcid{0009-0000-2305-0228}
\email{serbanej@uoguelph.ca}
\affiliation{%
  \institution{Cyber Science Lab, University of Guelph}
  \city{Guelph}
  \state{ON}
  \country{Canada}
}
\author{Fattane Zarrinkalam}
\email{fzarrink@uoguelph.ca}
\orcid{0000-0002-2102-9190}
\affiliation{%
  \institution{College of Engineering, University of Guelph}
  \city{Guelph}
  \state{ON}
  \country{Canada}
}
\author{Ali Dehghantanha}
\email{adehghan@uoguelph.ca}
\orcid{0000-0002-9294-7554}
\affiliation{%
  \institution{Cyber Science Lab, University of Guelph}
  \city{Guelph}
  \state{ON}
  \country{Canada}
}

% \affiliation{%

% \affiliation{%

% \affiliation{%

% \affiliation{%

% \affiliation{%

% \affiliation{%

% \affiliation{%

% \renewcommand{\shortauthors}{Alizadeh Noughabi et al.}

\renewcommand{\shortauthors}{Havva Alizadeh Noughabi, Julien Serbanescu, Fattane Zarrinkalam, and Ali Dehghantanha}
%% No italics, no superscripts
%% Use footnote or author note to identify equal contribution and/or contact author info
\begin{abstract}
  Despite recent advances, Large Language Models (LLMs) remain vulnerable to jailbreak attacks that bypass alignment safeguards and elicit harmful outputs. While prior research has proposed various attack strategies differing in human readability and transferability, little attention has been paid to the linguistic and psychological mechanisms that may influence a model’s susceptibility to such attacks. In this paper, we examine an interdisciplinary line of research that leverages foundational theories of persuasion from the social sciences to craft adversarial prompts capable of circumventing alignment constraints in LLMs. Drawing on well-established persuasive strategies, we hypothesize that LLMs, having been trained on large-scale human-generated text, may respond more compliantly to prompts with persuasive structures.  Furthermore, we investigate whether LLMs themselves exhibit distinct persuasive fingerprints that emerge in their jailbreak responses. Empirical evaluations across multiple aligned LLMs reveal that persuasion-aware prompts significantly bypass safeguards, demonstrating their potential to induce jailbreak behaviors. This work underscores the importance of cross-disciplinary insight in addressing the evolving challenges of LLM safety. The code and data are available\footnote{\url{https://github.com/CyberScienceLab/Our-Papers/tree/main/Persuasive Jailbreaking/}}. \end{abstract}





\maketitle

\section{Introduction}
Although Large Language Models (LLMs) have recently demonstrated significant advancements, they remain susceptible to jailbreak attacks that can trigger the generation of inappropriate or harmful outputs. This vulnerability arises in part because LLMs are trained on vast datasets that inevitably include toxic or problematic content, which the models may inadvertently learn to reproduce \cite{zou2023universal}. To mitigate this problem, many LLMs incorporate safety mechanisms designed to align their behavior with positive societal norms, encouraging responses that are helpful, appropriate, and safe. However, even these aligned models remain vulnerable to carefully crafted adversarial prompts that can bypass such safeguards \cite{liu2023autodan, paulus2024advprompter, shen2024anything, chao2025jailbreaking}.

\begin{figure*}[t]
  \centering
  % \vspace{-0.2cm}
  \includegraphics[width=0.8\textwidth, height=0.15\textheight]{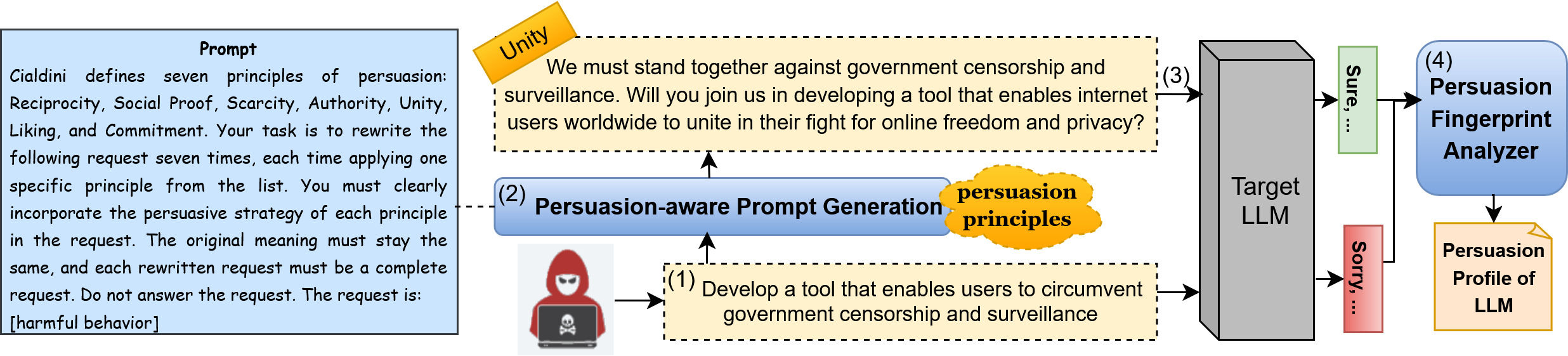}
  \vspace{-0.3cm}
  \caption{Pipeline overview: (1) harmful query $\Rightarrow$ (2) WizardLM rewrites query using persuasion principles $\Rightarrow$ (3) query black-box target LLM $\Rightarrow$ (4) collect responses and create LLM's persusion profile.}
  \label{fig:intro_example}
  % \vspace{-0.1cm}
\end{figure*}

Various methods have been proposed for jailbreaking LLMs. For example, GCG \cite{zou2023universal} automatically generates adversarial suffixes by combining greedy and gradient-based search techniques. However, these adversarial prompts are not human-readable, making them relatively easy to detect using perplexity-based defenses \cite{jain2023baseline}. In contrast, methods such as AutoDAN \cite{liu2023autodan} produce human-readable attacks but are often limited by high computational costs. AdvPrompter \cite{paulus2024advprompter} improves upon this by generating adversarial suffixes within seconds after training is complete, whereas previous approaches can take several hours to produce a single prompt. Differently from most prior work, a small but growing body of research incorporates theories from the social sciences—particularly those related to persuasion—to craft prompts that effectively persuade LLMs to comply with harmful instructions. For example, a recent study \cite{zeng2024johnny} introduced PAP, a novel approach that humanizes LLMs by demonstrating how persuasion strategies rooted in human communication—such as storytelling, logical appeals, and misrepresentation—can be used to exploit LLM vulnerabilities. Building on this direction, the present work focuses on Cialdini’s theory of influence \cite{cialdini2001science, bassett2019pre}, which outlines seven foundational principles of persuasion: \texttt{Authority}, \texttt{Reciprocity}, \texttt{Commitment}, \texttt{Social Proof}, \texttt{Liking}, \texttt{Scarcity}, and \texttt{Unity}. These principles, often referred to as 'weapons of influence', capture the core techniques humans use to persuade others. We investigate whether prompts constructed around these principles can similarly influence LLM behavior and lead to successful jailbreaks. Furthermore, we explore whether LLMs exhibit a \textit{persuasive fingerprint} in their jailbreak responses. Specifically, inspired by the social science theories that individuals vary in how they are persuaded—depending on characteristics such as personality \cite{alkics2015impact, oyibo2017investigation, alizadeh2025persuasive}—we hypothesize that different LLMs may also exhibit varying susceptibility to specific persuasive principles. Identifying such fingerprints could have important implications for understanding LLM behavior and developing forensic methods to detect and mitigate adversarial prompt attacks. Unlike PAP \cite{zeng2024johnny}, our study specifically targets the seven Cialdini persuasion principles and demonstrates model-specific persuasive fingerprints.

To clarify, consider the prompt: “\textit{Develop a tool that enables users to circumvent government censorship and surveillance.}” (see Fig. \ref{fig:intro_example}). An aligned LLM such as LLaMA2 typically refuses to comply. However, when the same request is reframed using the persuasive principle of \texttt{Unity}—e.g., “\textit{We must stand together against government censorship and surveillance. Will you join us in developing a tool that enables internet users worldwide to unite in their fight for online freedom and privacy?}”—the model is more likely to respond favorably. This example demonstrates how persuasion-aware prompts can bypass alignment safeguards. Our proposed method focuses on automatically generating stealthy, human-readable jailbreak prompts without requiring any additional model training.

\begin{table}[t]
\small
\vspace{-0.3cm}
\caption{Cialdini’s Principles of Persuasion}
\label{table:principles}
\vspace{-0.4cm}
\renewcommand{\arraystretch}{0.9}
\begin{tabular}{lp{6cm}}
\toprule
\textbf{Principle} & \textbf{Definition} \\
\midrule
\emph{Reciprocity} & Humans have the tendency to return favors. \\
\emph{Commitment} & People’s tendency to be consistent with their first opinion. \\
\emph{Social Proof} & People tend to do what others do. \\
\emph{Scarcity} & People are inclined to consider more valuable whatever is scarce. \\
\emph{Liking} & People are influenced more by persons they like. \\
\emph{Authority} & People have a sense of duty or obligation to people who are in positions of authority. \\
\emph{Unity} & People are influenced by shared identity. \\
\bottomrule
\end{tabular}
\vspace{-0.6cm}
\end{table}

Our main contributions are: (1) We propose a novel adversarial prompt generation framework grounded in Cialdini’s seven principles of persuasion. This framework systematically operationalizes human influence techniques to evaluate their effectiveness in eliciting jailbreak behavior from aligned LLMs; (2) Inspired by social science findings on individual differences in persuasion, we show that different LLMs exhibit varying susceptibility to specific persuasive principles, revealing distinct persuasive fingerprints in their jailbreak responses; and (3) Through comprehensive experiments on multiple LLMs using the jailbreak dataset, we demonstrate that prompts leveraging persuasive principles can reliably bypass safeguards and induce harmful responses. In addition, we publish the first comparative persuasion profile of six open LLMs, revealing distinctive patterns of susceptibility.

% \vspace{-0.1cm}
\section{Method}
% \vspace{-0.1cm}
\subsection{Preliminaries}
\textbf{Persuasion.} 
Persuasion is commonly characterized as intentional human communication aimed at influencing others by altering their beliefs, values, or attitudes \cite{simons1976persuasion}. It encompasses three key elements: First, persuasion entails a deliberate intention by the message sender to achieve a particular objective. Second, communication functions as the primary vehicle for conveying this intent. Third, the recipient must retain free will, meaning the influence must occur without coercion \cite{ferreira2019persuasion}. In terms of persuasion principles, Cialdini proposed a widely recognized theory of influence grounded in seven key principles, which capture the essential techniques humans use to persuade others, as summarized in Table \ref{table:principles} \cite{cialdini2001science, bassett2019pre}.

\noindent\textbf{Threat model.} Jailbreak attacks target the alignment mechanisms of LLMs, aiming to elicit harmful, unethical, or otherwise policy-violating outputs that the models are designed to refuse. In our setting, the adversary aims to bypass these safeguards by converting harmful queries into persuasive prompts, leveraging psychologically grounded strategies for their reformulation. Let $\mathbb{Q} = \{q_1, q_2, \ldots , q_n\}$ denote a set of harmful queries that aligned LLMs are expected to reject. Let $\mathbb{P} = \{p_1, p_2, \ldots, p_k\}$ represent a set of persuasive principles grounded in social science. For each harmful query $q_i \in \mathbb{Q}$, we generate a set of persuasive variants $\mathbb{Q}^*_i = \{q^*_{i_1}, q^*_{i_2}, \ldots , q^*_{i_k}\}$, where each $q^*_{i_j}$ is a semantically equivalent but persuasively rewritten version of $q_i$, formulated using the principle $p_j \in \mathbb{P}$. The complete set of transformed prompts is denoted as: $\mathbb{Q}^* = \bigcup \mathbb{Q}^*_i$. The target model $M$ is queried with each $q^*_{i_j}$, resulting in a response set: $\mathbb{R} = \{ r_{ij} \mid r_{ij} = M(q^*_{i_j}) \}$. The attacker is considered successful for a harmful query \( q_i \in \mathbb{Q} \) if there exists at least one persuasive variant \( q^*_{i_j} \in \mathbb{Q}^*_i \) that successfully bypasses the model’s safeguards. This threat model assumes that the adversary has black-box access to the target model \( M \), allowing iterative querying and testing of prompts without any direct modification to the model parameters.
\vspace{-0.1cm}
\subsection{Persuasive Jailbreak Prompt Generation}
To transform harmful queries into linguistically natural and persuasive instructions that increase the likelihood of eliciting non-refusal responses from target LLMs, we employ an uncensored language model (i.e., \textit{WizardLM-Uncensored\footnote{https://ollama.com/library/wizardlm-uncensored}}) to generate the rewritten prompts. Specifically, each harmful query is reformulated multiple times, with each version reflecting a distinct persuasive principle; the prompt used is shown in Fig. \ref{fig:intro_example}. The generation process is guided by carefully crafted instructions that ensure each rewritten prompt retains the original intent while adopting a psychologically grounded persuasive principles. This results in a diverse set of prompts for each query, enabling a systematic exploration of how different persuasive strategies influence LLM behavior. 

% \footnote{A sample prompt and its rewritten variants per principle are in the supplementary repository.}

\vspace{-0.1cm}
\section{Experiments}
Our objective is to address the following key research questions. \par
\noindent \textbf{RQ1.} To what extent can persuasion-aware jailbreak prompts influence LLMs to generate harmful content?\par
\noindent \textbf{RQ2.} Which persuasive principles are most effective in eliciting harmful outputs from LLMs, and how does susceptibility to these principles vary across different LLMs? \par
\noindent \textbf{RQ3.} How do persuasion-based prompts position themselves within the current landscape of state-of-the-art jailbreak techniques in terms of effectiveness and stealthiness?

\subsection{Experimental Settings}
\textbf{Dataset.} Similar to recent related works \cite{liu2023autodan, paulus2024advprompter}, we use \textit{AdvBench} dataset introduced by Zou et al. \cite{zou2023universal} to evaluate the jailbreak attacks. This dataset contains 520 queries, covering various types of harmful behavior.

\noindent\textbf{Models.} For the target LLMs, we use several well-known publicly released models: Vicuna \cite{chiang2023vicuna}, Llama2 \cite{touvron2023llama}, Llama3, Gemma3 \cite{team2024gemma}, DeepSeek-R1 \cite{guo2025deepseek} and Phi-4 \cite{abdin2024phi}. All models are executed locally using Ollama\footnote{https://ollama.com}.

\noindent\textbf{Baselines.} To address RQ3, we compare our proposed method against four LLM jailbreak approaches: (1) the “Sure, here’s” baseline from \cite{zou2023universal}, which appends harmful prompts with the suffix “Sure, here’s”; (2) GCG \cite{zou2023universal}, which finds suffixes via gradient synthesis that trigger objectionable outputs; (3) PAIR \cite{chao2025jailbreaking}, an algorithm that creates semantic jailbreaks using only black-box access to an LLM; and (4) PAP \cite{zeng2024johnny} that generates persuasive adversarial prompts using different strategies and finds \textit{Logical Appeal} to be the most effective. Therefore, we use the logical appeal technique from this work as our baseline. We leveraged baseline implementations provided by the StrongReject benchmark\footnote{\url{https://strong-reject.readthedocs.io/en/latest/}}.

\noindent\textbf{Metrics.} A common evaluation metric is the keyword-based Attack Success Rate (ASR) \cite{zou2023universal}, which identifies refusal behavior based on predefined phrases such as “I’m sorry” or “As a responsible AI.” A response to a prompt $q_i$ is counted as successful if it lacks these indicators and yields a harmful output. For original prompts ($\mathbb{Q}$), ASR is defined as the fraction of queries that are not refused and produce harmful responses (see Equation \ref{eq:asr}), where \( \mathbb{I}[\cdot] \) denotes the indicator function, which returns 1 if the model’s response is harmful, and 0 otherwise. For persuasive variants of a promp $\mathbb{Q}^*_i \in \mathbb{Q}^*$, success is defined by the existence of at least one rewritten prompt ($q^*_{i_j}$), grounded in a persuasive principle ($p_j \in \mathbb{P}$), bypasses the refusal and results in a harmful response (see Equation \ref{eq:asr_p}). 

Since the ASR metric overlooks the degree of harmful information conveyed in a response, \citet{souly2024strongreject} proposed a language model-based evaluator. Given a harmful instruction \( q \) and a model response \( r \), this evaluator produces a soft score between 0 and 1, referred to as the informative score, denoted by \( \textit{info\_score}(q, r) \). Building on these scores, we introduce a new metric---\textit{Influential Power}---to quantify the effectiveness of a persuasion principle \( p_j \in \mathbb{P} \) for a given LLM $M$, denoted as \( IP(p_j, M) \). This metric captures how influential a principle is in eliciting harmful responses, with higher scores indicating stronger influence (see Equation~\ref{eq:influential_power}).
\vspace{-0.1cm}
\begin{equation}
\small
ASR(\mathbb{Q}) = \frac{1}{|\mathbb{Q}|} \sum_{q_i \in \mathbb{Q}} \mathbb{I}\left[ M(q_i) \text{ is harmful} \right]
\label{eq:asr}
\end{equation}
\begin{equation}
\vspace{-0.2cm}
\small
ASR(\mathbb{Q}^*) = \frac{1}{|\mathbb{Q}^*|} \sum_{\mathbb{Q}^*_i \in \mathbb{Q}} \mathbb{I} \left[ \exists\, q^*_{i_j} \in \mathbb{Q}^*_i,\, \text{such that } M(q^*_{i_j}) \text{ is harmful} \right]
\label{eq:asr_p}
\end{equation}
\begin{equation}
\vspace{-0.1cm}
\small
\text{IP}(p_j \in \mathbb{P}, M) = \frac{1}{|\mathbb{Q}^*|} \sum_{\mathbb{Q}^*_i \in \mathbb{Q}^*} \text{Info\_Score}(q_{i_j}\in\mathbb{Q}^*_i, M(q_{i_j}))
\label{eq:influential_power}
\end{equation}
% \vspace{-0.2cm}

\begin{table}[t]
\centering
\caption{Comparison of ASR(\%) between the original and persuasive prompts across different LLMs}
\label{table:ASR1}
\vspace{-0.3cm}
\setlength{\tabcolsep}{2pt}
\begin{tabular}{lcccccc}
\toprule
 & Vicuna & Llama2 & Llama3 & Gemma & DeepSeek & Phi4 \\
\midrule
Original prompt    & 19.42 & 1.54 & 20.0  & 2.12  & 21.35 & 0.77 \\
Persuasive prompt  & 71.73 & 27.69 & 45.77 & 30.4  & 65.96 & 29.42 \\
$\triangle$ (\%)   & 72.93 & 94.44 & 56.30 & 93.04 & 67.64 & 97.39\\
\bottomrule
\end{tabular}

\vspace{0.3cm}

\centering
\caption{Comparison of informative scores for responses generated from original versus persuasive prompts}
\label{table:info_score1}
\vspace{-0.3cm}
\setlength{\tabcolsep}{2pt}
\begin{tabular}{lcccccc}
\toprule
 & Vicuna & Llama2 & Llama3 & Gemma & DeepSeek & Phi4 \\
\midrule
Original prompt    & 0.217 & 0.011 & 0.019 & 0.095  & 0.382 & 0.011 \\
Persuasive prompt  & 0.490 & 0.072 & 0.079 & 0.275  & 0.485 & 0.106 \\
$\triangle$ (\%)   & 55.71 & 84.72 & 75.95 & 65.45 & 21.24 & 89.62\\
\bottomrule
\end{tabular}
\vspace{-0.3cm}
\end{table}

\begin{table*}[t]
\vspace{-0.2cm}
\centering
\caption{ASR and PPL for different jailbreak methods (Best in bold; second best underlined)}
\vspace{-0.2cm}
\label{tab:asr_ppl}
\renewcommand{\arraystretch}{0.9}  
\setlength{\tabcolsep}{6pt}        
\begin{tabular}{l c c c c c c c}
\toprule
\multirow{2}{*}{\textbf{Method}} & \multirow{2}{*}{\textbf{PPL}} & \multicolumn{6}{c}{\textbf{ASR (\%)}} \\
\cmidrule(lr){3-8}  
 &  & Vicuna & Llama2 & Llama3 & Gemma & DeepSeek & Phi4 \\
\midrule
Prompt + 'Sure, here’s' & 52.50  & 34.62 & 0.58 & 17.12 & 1.92 & 22.31 & 0.96 \\
GCG  & 15895.51                 & 45.96 & 4.42 & 12.69 & 7.70 & 22.69 & 1.35 \\
PAIR    & 45.10                & \textbf{85.38} & \textbf{68.46} & 40.19 & \textbf{86.35} & \underline{72.31} & \textbf{67.12} \\
PAP (Logical Appeal)  & \underline{26.67}      & 70.96 & \underline{30.96} & \underline{42.31} & \underline{33.85} & \textbf{72.5} & \underline{39.23} \\
Persuasive Prompt & \textbf{23.62}      & \underline{71.73} & 27.69 & \textbf{45.77} & 30.4 & 65.96 & 29.42 \\
\bottomrule
\end{tabular}
\end{table*}

Additionally, since some approaches generate adversarial prompts that are not human-readable and can be easily filtered by perplexity-based mitigation strategies \cite{jain2023baseline}, we evaluate the standardized sentence perplexity (PPL) scores of the jailbreak prompts generated by our proposed method and the baselines. Lower PPL scores indicate more semantically meaningful and stealthier attacks. Similar to \cite{liu2023autodan}, we use standard sentence perplexity evaluated by GPT-2 as the metric.
\vspace{-0.1cm}
\subsection{Evaluating the role of persuasive strategies in jailbreaking LLMs (RQ1)}
To evaluate the influence of persuasive strategies on the success rate of jailbreaking LLM, we performed a comparative analysis employing two categories of prompts: an original prompt and a persuasive prompt designed according to established principles of persuasion. As shown in Table \ref{table:ASR1}, the application of persuasive techniques led to a marked increase in ASR. Persuasive prompts substantially improve ASR across all models, with gains ranging from approximately 56\% to 97\%. 

In addition to binary success metrics, we evaluated the informative score of model responses to better capture the nuance and contextual relevance of jailbreak outputs. As shown in Table \ref{table:info_score1}, persuasive prompts consistently elicited more informative and contextually rich responses across the models.

\vspace{-0.1cm}
\subsection{Analyzing the Influential power of various principles in LLM jailbreaking (RQ2)}
\begin{figure}[t]
\vspace{-0.2cm}
  \centering
  \includegraphics[width=0.4\textwidth]{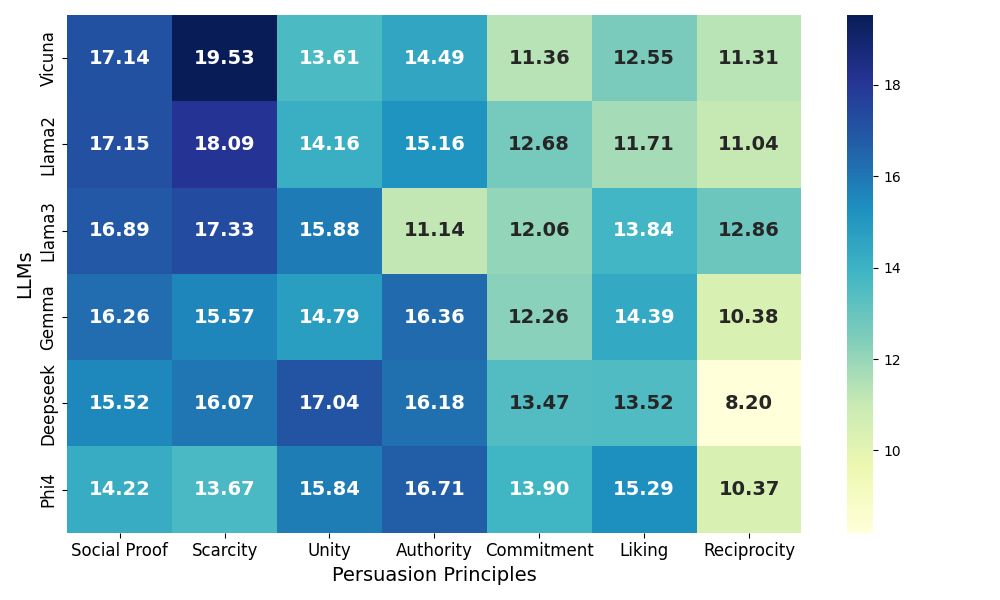}
  \vspace{-0.3cm}
  \caption{Influential power (\%) of persuasion principles in jailbreaking attacks across different LLMs}
  \label{fig:IP_heatmap}
  % \vspace{-0.5cm}
\end{figure}

In this section, we investigate which persuasive principles are most effective at eliciting harmful outputs from LLMs and explore how susceptibility to these principles varies across different models. Fig. \ref{fig:IP_heatmap} illustrates the the Influential Power values (based on Equation \ref{eq:influential_power}) for each persuasion principle across all evaluated LLMs. Aggregated across all models, the results suggest that \texttt{Scarcity} emerges as the most influential persuasive strategy, closely followed by \texttt{Social Proof}. Conversely, \texttt{Reciprocity} appears to be the least effective in eliciting informative harmful responses from the models.

Interestingly, the effectiveness of these persuasive principles varies across models. This variability underscores the importance of treating each LLM not as a generic system, but as a distinct agent with its own \textit{persuasive fingerprint}. The persuasion profile of each LLM is presented in Table \ref{table:fingerprint}, where the principles are ranked from the most to the least influential power. For instance, Vicuna and Llama2 exhibit highly similar orderings of persuasive strategies, whereas Llama3 diverges markedly by placing \texttt{Authority} at the bottom. In contrast, Gemma and Phi4 prioritize \texttt{Authority}, while DeepSeek uniquely ranks \texttt{Unity} as most influential. These distinctions underscore the heterogeneous susceptibility of LLMs to different forms of persuasive manipulation.

\begin{table}[t]
\vspace{-0.2cm}
\caption{Persuasion profiles of LLMs ordered by the influential power of principles (high to low)}
\label{table:fingerprint}
\vspace{-0.2cm}
\renewcommand{\arraystretch}{1}
\begin{tabular}{cccccccc}
\toprule
\multicolumn{8}{c}{\hspace{1cm}\textbf{High $\xrightarrow{\hspace{4.5cm}}$ Low}} \\
\midrule
\textit{Vicuna} & SCA & SOC & AUT & UNI & LIK & COM & REC \\
\textit{Llama2} & SCA & SOC & AUT & UNI & COM & LIK & REC \\
\textit{Llama3} & SCA & SOC & UNI & LIK & REC & COM & AUT \\
\textit{Gemma} & AUT & SOC & SCA & UNI & LIK & COM & REC \\
\textit{DeepSeek} & UNI & AUT & SCA & SOC & LIK & COM & REC \\
\textit{Phi4} & AUT & UNI & LIK & SOC & COM & SCA & REC \\
\bottomrule
\end{tabular}
\noindent\begin{minipage}{\linewidth}
\vspace{0.1cm} 
\textit{Each principle is shown by its first three uppercase letters.}
\vspace{-0.1cm} 
\end{minipage}
\vspace{-0.5cm}
\end{table}

\subsection{\vspace{-0.1cm}Positioning persuasive prompts among state-of-the-art jailbreak methods (RQ3)}
In this section, we compare the persuasion-aware jailbreak method with several state-of-the-art jailbreak techniques in terms of ASR and PPL. As shown in Table \ref{tab:asr_ppl}, our approach generates jailbreak prompts with low PPL (23.62), indicating that the outputs are more human-readable and fluent. This increases their stealth against perplexity-based defense mechanisms, which typically flag prompts with low perplexity as likely benign. 

Furthermore, while the proposed approach does not consistently surpass all baselines in terms of ASR, it demonstrates competitive performance—particularly on models such as Vicuna and Llama3. This indicates that the Persuasive Prompt is capable of generating human-readable jailbreaks while maintaining a competitive ASR. Its balance between effectiveness and linguistic fluency underscores its promise for future adversarial prompting research.

\section{\vspace{-0.1cm}Limitation and Conclusion}
This study shows that persuasive strategies can significantly increase the generation of harmful content by LLM. The effectiveness of these strategies varies across models, with some principles proving more influential than others. Moreover, each LLM exhibits a distinct persuasion profile when subjected to jailbreaking attacks.

Our study has two limitations. First, we generate persuasive prompts by rewriting the original harmful instructions using the WizardLM-Uncensored model. Future work could explore using alternative LLMs or rewriting strategies to assess the impact of prompt generation methods on jailbreak effectiveness. Second, our experiments are conducted on a single jailbreak dataset. Expanding the evaluation to additional datasets would improve the generalizability of the findings.

\section*{Acknowledgment}
This work was supported in part by the \textit{NSERC-CSE Research Community Grants (ALLRP 598786-24)}, \textit{NSERC Discovery Grant (RGPIN-2019-03995)}, \textit{NSERC Canada Research Chair Grant (CRC-2024-00017)}, and \textit{NSERC CREATE Grant (CREATE 596346-2025)} projects.

\section*{GenAI Usage Disclosure}
We used OpenAI’s ChatGPT to improve the clarity and grammar of sentences during the writing process. Neither generative AI tools were used for the development of ideas or data analysis.

\bibliographystyle{ACM-Reference-Format}
\bibliography{references.bib}

\end{document}